\PassOptionsToPackage{dvipsnames}{xcolor}

\documentclass{article}
\usepackage[utf8]{inputenc} 
\usepackage[T1]{fontenc}    
\usepackage{hyperref}       
\usepackage{url}            
\usepackage{booktabs}       
\usepackage{amsfonts}       
\usepackage{nicefrac}       
\usepackage{microtype}      
\usepackage{threeparttable}
\usepackage{wrapfig}
\usepackage{natbib}
\usepackage{xcolor}
\usepackage{amsmath}
\usepackage{xcolor}
\usepackage{arxiv}
\usepackage{url}
\newcommand\myshade{85}
\colorlet{mylinkcolor}{black}
\colorlet{mycitecolor}{violet}
\colorlet{myurlcolor}{YellowOrange}

\hypersetup{
  linkcolor  = mylinkcolor!\myshade!black,
  citecolor  = mycitecolor!\myshade!black,
  urlcolor   = myurlcolor!\myshade!black,
  colorlinks = true,
}

\usepackage{color}
\usepackage{times}
\usepackage{latexsym}
\usepackage[T1]{fontenc}
\usepackage[utf8]{inputenc}
\usepackage{microtype}
\usepackage{booktabs}
\usepackage{graphicx}
\usepackage{multirow}
\usepackage{float}
\usepackage[frozencache, cachedir=minted-cache]{minted}
\definecolor{navyblue}{RGB}{0,0,128}

\title{
Pixel Sentence Representation Learning
}

\author{
    \textbf{Chenghao Xiao}\textsuperscript{1}\thanks{Equal contribution.}\quad 
    \textbf{Zhuoxu Huang}\textsuperscript{2}\footnotemark[1]\quad 
   \textbf{ Danlu Chen}\textsuperscript{3}\quad \\
    \textbf{G Thomas Hudson}\textsuperscript{1}\quad 
    \textbf{Yizhi Li}\textsuperscript{4}\quad 
    \textbf{Haoran Duan}\textsuperscript{1}\quad \\
    \textbf{Chenghua Lin}\textsuperscript{4}\quad
    \textbf{Jie Fu}\textsuperscript{6}\quad
    \textbf{Jungong Han}\textsuperscript{5}\quad
    \textbf{Noura Al Moubayed}\textsuperscript{1}\quad
\\
\small
    \textsuperscript{1}Durham University\quad \textsuperscript{2}Aberystwyth University \quad 
    \textsuperscript{3}UC San Diego\\
\small
    \textsuperscript{4}University of Manchester\quad 
    \textsuperscript{5}University of Sheffield\quad
    \textsuperscript{6}Hong Kong University of Science and Technology \quad 
\\
\small
\texttt{chenghao.xiao@durham.ac.uk} \quad 
\texttt{} \quad
}

\begin{document}

\maketitle

\begin{abstract}
Pretrained language models are long known to be subpar in capturing sentence and document-level semantics. Though heavily investigated, transferring perturbation-based methods from unsupervised visual representation learning to NLP remains an unsolved problem.
This is largely due to the discreteness of subword units brought by tokenization of language models, limiting small perturbations of inputs to form semantics-preserved positive pairs. In this work, we conceptualize the learning of sentence-level textual semantics as a visual representation learning process. Drawing from cognitive and linguistic sciences, we introduce an unsupervised visual sentence representation learning framework, employing visually-grounded text perturbation methods like typos and word order shuffling, resonating with human cognitive patterns, and enabling perturbation to texts to be perceived as continuous. Our approach is further bolstered by large-scale unsupervised topical alignment training and natural language inference supervision, achieving comparable performance in semantic textual similarity (STS) to existing state-of-the-art NLP methods. 
Additionally, we unveil our method's inherent zero-shot cross-lingual transferability and a unique leapfrogging pattern across languages during iterative training. To our knowledge, this is the first representation learning method devoid of traditional language models for understanding sentence and document semantics, marking a stride closer to human-like textual comprehension.
Our code is available at \url{https://github.com/gowitheflow-1998/Pixel-Linguist}
\end{abstract}

\section{Introduction}
\label{sec:intro}
Vanilla language models are long known to have subpar sentence-level representation \citep{reimers2019sentence,wang2023improving}, even worse than averaging static word embeddings \citep{pennington2014glove}, i.e., sentence representations attained by pooling from sub-word embeddings encoded by language models do not closely reflect the relative semantics of sentences. Encouraged by the remarkable success of visual representation learning facilitated by unsupervised contrastive learning \citep{chen2020simple,he2020momentum}, efforts in NLP are made to leverage unsupervised contrastive learning to recover sentence-level encoding abilities from the models \citep{fang2020cert,wu2020clear,gao2021simcse,meng2021coco}.

However, translating the advancements in visual representation learning to learning sentence-level textual semantics presents unique challenges: a single augmentation \citep{wu2020clear,meng2021coco} might alter the meaning of a sentence, posing problems of the validity of the augmented sentence as a positive pair. Such attempts are primarily bottlenecked by the discreteness of subword units brought by tokenization \citep{sennrich2016neural}, impeding the creation of continuous unsupervised semantic pairs that have preserved semantics through small perturbations to inputs. 
\begin{figure}
    \centering
    \includegraphics[width=0.65\linewidth]{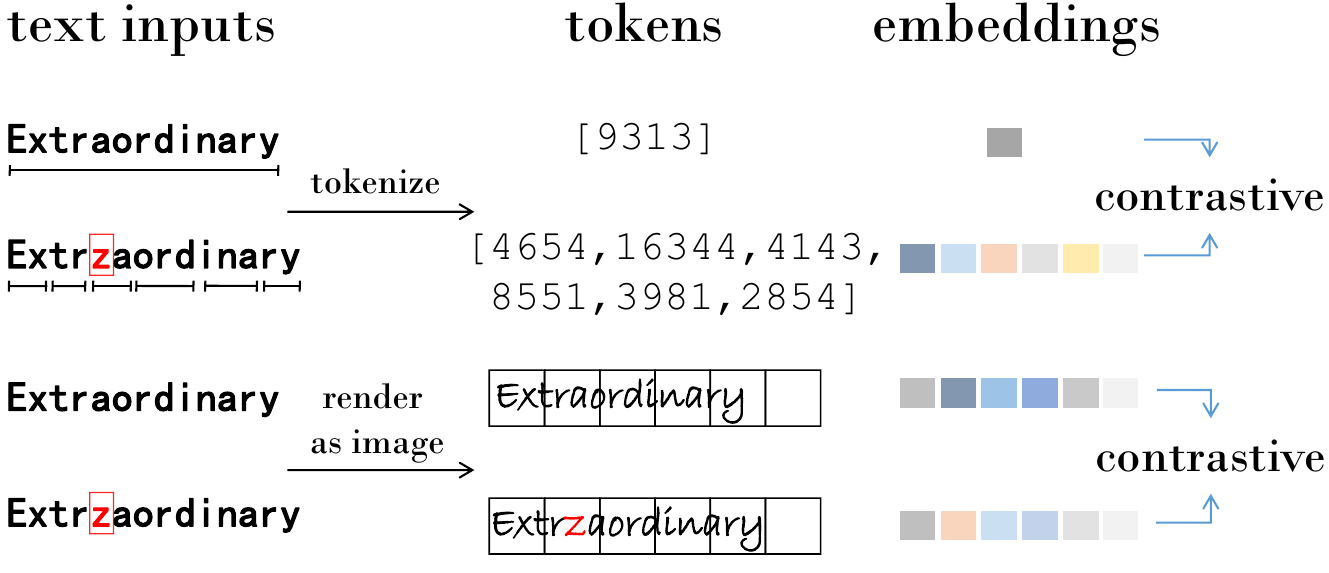}
    \caption{Perceptual difference between tokenization-based language models and vision models, with the example of the word  ``extraordinary'' with one single typo injected.}

    \label{fig:figure1}
\end{figure}

Therefore, the most recognized unsupervised sentence representation learning method in NLP applies two dropout masks to the identical input to attain two representations, as positive pairs in contrastive learning~\citep{gao2021simcse}. 

We argue that using identical inputs confines the method of \citet{gao2021simcse} to essentially only a way to improve uniformity \citep{wang2020understanding} by distancing negative examples that are not identical to an instance itself, lacking the capability to provide signals towards non-verbatim examples, which are crucial for capturing semantically similar sentences.

Figure~\ref{fig:figure1} encapsulates the difference between tokenization-based language models and vision models, on the perception of text. Using \texttt{bert-base-uncased} tokenizer \citep{devlin2018bert}, the word \texttt{extraordinary} is a standalone token [9313], while with one single typo \texttt{z} injected, \texttt{extrzaordinary} is tokenized into [4654, 16344, 4143, 8551, 3981, 2854], causing a large perceptual shift for the model. This mechanism has hindered traditional language models in recognizing that these minor textual perturbations do not fundamentally alter the underlying semantics. 

On the other hand, the inherent continuity in visual models grants them less perceptual variance for textual perturbations. In parallel, we recognize that human understanding of text is not only visually grounded, but also tolerant of irregularities such as typos, varied word orders, and distorted text presentations \citep{rawlinson1976,ferreira2007good,grainger2011dual,rayner2012psychology}. 

Motivated by these, we propose a novel ``pixel sentence representation learning'' framework that \textbf{redefines the learning of sentence and document-level textual semantics as a visual representation learning process}, taking the perceptually continuous advantages of vision models and closely mirroring the human cognitive processes. The unique approach diverges from traditional tokenization-based language models, allowing the models to effectively leverage the rich multi-modal semantic signals inherent in text, providing an alternative avenue for achieving a more natural understanding of textual semantics. Importantly, we also introduce the superiority of modeling sentence semantics in the pixel space to extrapolate and generalize to semantics of un-trained languages.

The main contributions of this work are:

\begin{itemize}
    \item We present and validate the potential of learning sentence and document-level semantics as a visual representation learning process, and design a progressive alignment scheme to facilitate the framework.
    \item Inspired by cognitive and linguistic sciences, we utilize typos and word-order shuffling as visually-grounded unsupervised augmentation methods, overcoming the challenges of applying perturbation augmentation methods in NLP due to discreteness brought by tokenization.
    \item We uncover a surprising leapfrogging pattern in pixel-based language models through iteratively training on OOD cross-lingual pairs and revisiting English NLI, showcasing an epiphany-like advancement in semantic understanding by ``taking hints'' across languages.
    \item We train and open-source the Pixel Linguist model series\footnote{\url{https://huggingface.co/Pixel-Linguist/Pixel-Linguist-v0}}, providing the research community with an alternative avenue for achieving a more natural and intuitive understanding of textual semantics.
\end{itemize}

\section{On the Behavioral Gap between Language Models and Pixel Models}\label{sec:gap}

To our knowledge, we are the first to leverage pure vision models for learning sentence and document-level text representation. In this section, we take a conscious look at 1) the motivation of this approach, and 2) the status quo of available techniques to facilitate this novel idea. We conduct 3 proof-of-concept experiments to understand the behavioral differences between vanilla tokenization-based LMs and their pixel counterparts.

\subsection{Preliminary}
\paragraph{Pixel Sentence and Document-level Representation Learning} We define pixel sentence and document-level representation learning as the process of understanding sentence and document-level text using vision models. 
The representations of sentences or documents can be used to reflect relative semantic relationships with one another, ideally with simple similarity match without further projection, which approximates relative semantics drawn from the real-world distribution. Formally, given real-world pairwise data $i,j$ sampled from $p$\textsubscript{data}, we aim to minimize between the model's similarity perception of $i,j$, and their ground-truth relative semantics $s_{ij}$.
\begin{equation}
\min_{f \in \mathcal{F}} \mathop{\mathbb{E}}_{\stackrel{\text{i.i.d.}}{(i, j) \sim p_{\text{data}}}} \left[ \left( f(x_i)^\top f(x_j) - s_{ij} \right)^2 \right] + \lambda \sum_{i=1}^{N} \left \| f(x_i) \right \|_p^p,
\end{equation}
where $x_i$ is a text, $f \in \mathcal{F}$ is a vision model, making $f(x_i)$ a visual representation of text.

This has distinguished our work from other lines of work, including Image Representation Learning with vision encoders (e.g., MOCO, SimCLR \citep{he2020momentum,chen2020simple}), Sentence Representation Learning with text encoders (e.g., SBERT, SimCSE \citep{reimers2019sentence,gao2021simcse}), Image-Text Representation Learning with multi-modal encoders (e.g., CLIP \citep{radford2021learning}), and Image-Text Representation Learning with only vision encoders (e.g., CLIPPO \citep{tschannen2023clippo}). The only work that fully aligns with our scope (text understanding with vision encoders, without non-text image signals) is PIXEL \citep{rust2023language}. However, PIXEL is a general-purpose vanilla model like BERT \citep{devlin2018bert}, requiring fine-tuning to adapt to further downstream tasks. As we will show, the vanilla sentence and document-level representation provided by this backbone largely falls behind its NLP vanilla counterparts.

\subsection{Observation 1: Robustness to Text Perturbations} 

As one of our main cognitive inspirations, we measure the behaviors of vanilla tokenization-based language models \citep{devlin2018bert} and a pixel-based model \citep{rust2023language} under text perturbations. 

We perform each attack outlined in Table~\ref{tab: poc2} on all \texttt{sentence 1} in STS-b \citep{cer2017semeval}, and measure the embeddings' cosine distance shifted from the original emebeddings (\textit{Rep. Shift}), and the degradation of relative semantics performance when evaluating the attacked sentences 1 with original sentences 2 (\textit{Semantics}) on STS-b. Detailed descriptions of these attacks are given in Section~\ref{subsec: unsupervised methods}.

\begin{table}[hp]
\centering
\scalebox{0.76}{
\begin{tabular}{lccccc}\toprule
Model $\rightarrow$ & \multicolumn{2}{c}{BERT} & \multicolumn{2}{c}{PIXEL} \\
\cmidrule{2-5}
Perturbation $\downarrow$ & Rep. Shift & Semantics &Rep. Shift & Semantics \\
\midrule
\multicolumn{5}{l}{\textit{Charcter-level}}\\
Insertion & 0.049 & {\textcolor{red}{- 10.8$\downarrow$}} &0.049&\hspace{1.8mm}{\textcolor{navyblue}{2.6$\uparrow$}} \\
Deletion & 0.040 & {\textcolor{red}{\hspace{1mm} - 7.8$\downarrow$}}&0.048& {\textcolor{red}{- 0.4$\downarrow$}}\\
Substitution &0.049& {\textcolor{red}{- 10.3$\downarrow$}}&0.038&\hspace{2.9mm} {\textcolor{navyblue}{0.0$\leftrightarrow$}}\\
Neighbor Swap &0.047& {\textcolor{red}{- 11.6$\downarrow$}}&0.014&\hspace{2mm}{\textcolor{navyblue}{0.3$\uparrow$}}\\
\midrule
\multicolumn{5}{l}{\textit{Word-level}}\\
Random Shuffle &0.115&{\textcolor{red}{- 19.1$\downarrow$}}& 0.087 & \hspace{2.2mm}{\textcolor{navyblue}{4.6$\uparrow$}} \\
Condition Shuffle &0.079&{\textcolor{red}{- 17.6$\downarrow$}}&0.048&\hspace{2.2mm}{\textcolor{navyblue}{2.1$\uparrow$}}\\
\bottomrule
\end{tabular}
}
\vspace{.2cm}
\caption{Representation Distance Shift, and Sentence-level Semantics Change characterized by STS-b test performance.}
\label{tab: poc2}
\end{table}

Not surprisingly, due to its tokenization dependency, BERT degrades under character-level attacks. However, the greatest degradation occurs when word order is randomly shuffled, showing the non-trivial contribution of positional embeddings in vanilla BERT. Conversely, PIXEL shows less semantic sensitivity, and even surprisingly attains semantics gain on STS-b on 5 out of 6 perturbed methods evaluated. 

In conclusion, pixel-based language models' sensitivity of visually-grounded textual perturbations is orders of magnitude less (shown by character-level semantic shifts) than tokenization-based language models, and it is also less sensitive to positions of words (word-level semantic shifts). This behavior of pixel models has granted us the natural convenience of using perturbed examples in constructing unsupervised contrastive learning pairs - as they are already perceived similar before training, and thus not detrimental to the models as positive pairs.

\subsection{Observation 2: Potential for Zero-shot Cross-lingual Transferability}
\label{subset:poc3}
Pixel-based language models are tokenization-free and are thus ideal for cross-lingual transfer learning.
We adopt a representation degeneration perspective \citep{gao2018representation,ethayarajh2019contextual} to understand the zero-shot superiority of pixel language models in out-of-distribution (OOD) generalization. We measure the representation distribution of each language of the vanilla models using the multilingual sts-b (multilingual STS-b \citep{cer2017semeval,huggingface:dataset:stsb_multi_mt}), which spans 10 languages from 4 language families.

\vspace{-0pt}
\begin{table}[h]
\centering
\scalebox{0.88}{
\begin{tabular}{lrrrrr}
\toprule
 & en    & de    & nl    & es   & fr    \\
\midrule
BERT                 &  \textbf{0.763} & 0.895                           & 0.895                           & 0.901                           & 0.893 \\
PIXEL                & 0.833                           & \textbf{0.877} & \textbf{0.867} & \textbf{0.894} & \textbf{0.879} \\
\midrule
\midrule
 &it    & pt & pl    & ru    & zh    \\
\midrule
BERT                 & 0.900                           & 0.891                           & 0.914                           & 0.942                           & 0.909 \\
PIXEL                & \textbf{0.884} & \textbf{0.874} & \textbf{0.866} & \textbf{0.917} & \textbf{0.880} \\
\bottomrule
\end{tabular}
}
\vspace{.2cm}
\caption{Anisotropy Estimates ($\downarrow$ the better) of 10 languages. BERT and PIXEL are both trained on the same corpus with 4B tokens, whose major language is English.}
\label{tab: poc3}
\end{table}

The results presented in Table~\ref{tab: poc3} reveal key insights. We encode sentence-level embeddings from the test set of each language with mean-pooling, and estimate the anisotropy by calculating the empirical mean of pairwise cosine similarity among these embeddings.

While BERT presents a slightly more isotropic pattern on its in-distribution language (en), all OOD languages (i.e., not seen during pre-training) suffer from severe representation degeneration. The advantage of PIXEL is immediately pronounced in OOD languages, with isotropy levels surpassing BERT. The robustness of PIXEL in maintaining consistent representation distribution across diverse languages, suggests that its semantic understanding at the sentence level is not solely reliant on language-specific features. Instead, PIXEL appears to leverage a more universal, shape-based approach to semantic cognition, suggesting a natural cognitive alignment with humans.

As we further explore (Section~\ref{sec: cross-lingual transfer results}), when facilitated by contrastive learning, this alignment promises an amazingly strong bonding effect across languages, and provides a synergistic enhancement on unseen languages, evident in the model's zero-shot semantics understanding abilities.

\subsection{Dilemma: Unsatisfactory Semantics (yet)}

\begin{wraptable}{r}{0.45\textwidth}
\centering
\scalebox{0.93}{
\begin{tabular}{lccc}\toprule
Model &GloVe & BERT &PIXEL\\\midrule
\multicolumn{3}{l}{\textit{Sentence Semantics} ($\uparrow$)} \\
cls & - &\textbf{26.40} &21.20 \\
mean & \textbf{58.02} & 52.59 &16.28 \\
\bottomrule
\end{tabular}
}
\vspace{.2cm}
\caption{Sentence semantics of static word embeddings, vanilla language encoder and its pixel language encoder counterpart. 
}
\label{tab: poc1}
\end{wraptable}


Awareness has long been raised that vanilla LMs are worse at capturing sentence-level semantics than simply averaging static word embeddings \citep{reimers2019sentence,pennington2014glove}, with a plethora of research dedicated to overcoming this \citep{gao2021simcse,wang2023improving}.

By measuring STS-b \citep{cer2017semeval} performance, we show that, when pretrained on the same corpus with similar model architectures, the vanilla pixel counterpart \citep{rust2023language} of BERT presents a sentence-level semantics even significantly inferior to the already subpar performance of BERT \citep{devlin2018bert,reimers2019sentence}.

Despite their robustness to text perturbations and potential in zero-shot cross-lingual transferability, which we have shown previously, we reveal the dilemma that vanilla pixel models lag behind their LM counterparts in capturing sentence-level semantics. Therefore, our main contribution is to fill this gap and get the best of both worlds.

\section{Methods}

As discussed in Section \ref{sec:gap}, pixel-based language models are less sensitive to perturbations and present a lower representation degeneration on out-of-distribution languages. In this section, we present a novel pixel sentence representation learning framework designed based on these insights.

\begin{figure}[htbp]
    \centering
    \includegraphics[width=0.7\linewidth]{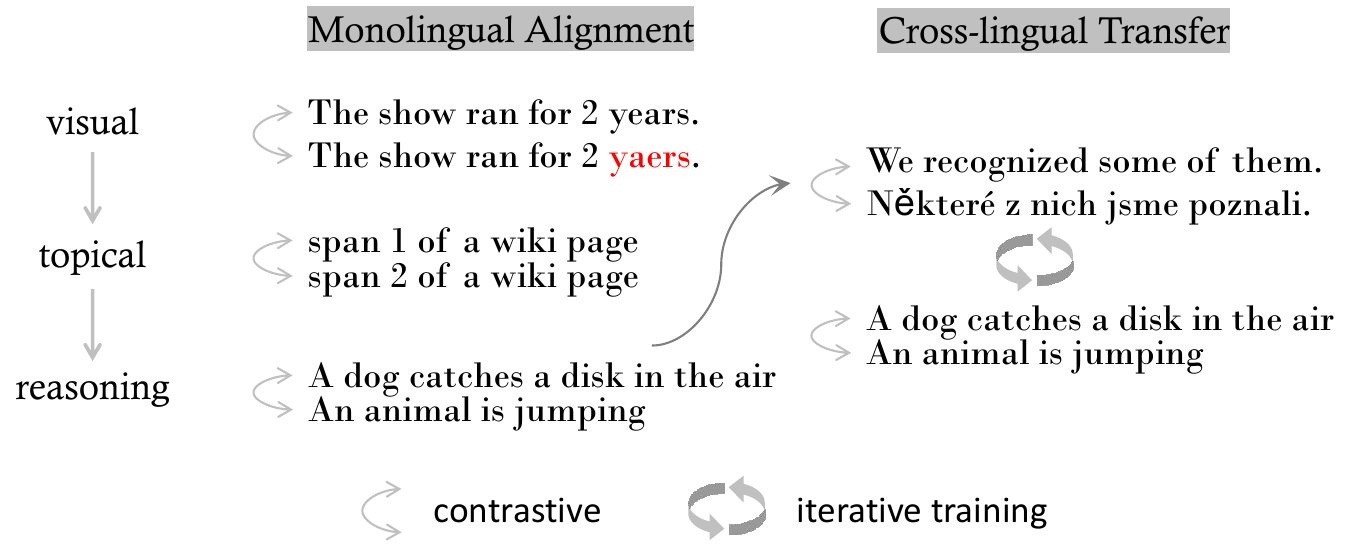}
    \caption{Training process.}
    \label{fig:training process}
\end{figure}

Our framework proposes 1) a progressive visual alignment (unsup.) - topical alignment (unsup.) - reasoning alignment (sup.) scheme for \textit{monolingual learning}, and 2) an iterative cycle training scheme of revisiting OOD language pairs and English natural language inference pairs, for \textit{multilingual learning}. We start by introducing the unsupervised and supervised learning methods that facilitate the framework.

\subsection{Unsupervised Learning}
\label{subsec: unsupervised methods}

Our unsupervised methods consist of a bag of visually-grounded language augmentation techniques, and a random-span sampling method for topical alignment. In the analysis section~\ref{ablation: tradition vis aug}, we also show that traditional visual augmentation methods do not work for the task.

\paragraph{Visually-grounded Language Augmentation} 
Our visually-grounded textual augmentation methods comprise of two classes of augmentation, concerning word order and typos, which are heavily inspired by linguistic cognitive studies and day-to-day experience that human reading is not sensitive to random word orders and typos \citep{rawlinson1976,ferreira2007good,grainger2011dual,rayner2012psychology}. The two methodologies and their intuitions are as follows.

\paragraph{(1) Word order shuffling} is an augmentation method inspired by the phenomenon that shuffled word orders do not affect making sense of the semantics much. For instance, in the sentence ``This a is computer vision research paper'', we as humans are able to make sense of the semantics, acting just as a bag-of-word model \citep{yuksekgonul2022and}.

Here, we design two operations, \textit{word random shuffling}, and \textit{word conditional shuffling}. a) For random shuffling, we shuffle all the words in the same document, allowing words to go across different sub-sentences. b) For conditional shuffling, we impose that the first and the last word of a sub-sentence must stay in the original position, while allowing the rest of the words to shuffle within the sub-sentence. 

\textbf{(2) Typos} serve as another source of augmentation which humans have tolerance towards. Leveraging resources from taxonomy of adversarial text attack in NLP \citep{li2019textbugger,morris2020textattack}, we select and incorporate four shape-based character-level typo attacks--\textit{Character Insertion}, \textit{Character Deletion}, \textit{Character Substitution}, and \textit{Character Neighbor Swap}--as augmentation methods. \textit{Character Deletion} randomly deletes a character in a sample (e.g., \texttt{sentence} $\rightarrow$ \texttt{setence}). \textit{Character Substitution} randomly replaces a character by either a character with similar visual appearance (e.g, \texttt{a} and \texttt{e}), or a character that is next to the original character on the keyboard (e.g, \texttt{a} and \texttt{s}). \textit{Character Neighbor Swap} randomly swaps the order of two adjacent characters (e.g., \texttt{random} $\rightarrow$ \texttt{radnom}). 

The superiority of word and character-level perturbation, when combining with vision encoders, compared to combining with text encoders, is the \textbf{continuity}. Were to process the perturbed texts with language models, a small perturbation would even lead to being tokenized into extremely different sub-words, depriving the useful signals providing to the language models (Recalling Figure~\ref{fig:figure1}).

Notably, to explore the limit of unsupervised language semantics acquisition solely dependent on visual learning, we only consider \textbf{visually-grounded text perturbation methods}. Therefore, even though there are many text augmentation methods, such as back translation \citep{edunov2018understanding}, and synonym replacement \citep{jin2020bert}, we do not consider them as ``unsupervised vision methods''.

\paragraph{Random Span Sampling} 
Following Visual Alignment, we facilitate Topical alignment with random span sampling.
We sample spans from the same document to form different views of the document. This method is inspired by unsupervised information retrieval learning \citep{lee2019latent,izacard2021unsupervised}. In this work, we conduct two independent span sampling to a document, allowing overlapping spans from the same document. Intuitively, the overlapping part encourages model's perception about lexical matching, providing with vision models a bridge to pick up the part from the two sequences that have a same shape, as a starting point to extrapolate the semantics of the remaining parts.

\subsection{Supervised Learning}

For supervised learning, we leverage paired language data. We use Natural Language Inference (NLI) datasets in the main experiments, including SNLI and MNLI. 

We then heavily explore cross-lingual zero-shot transfer, by collecting high-quality sentence pairs from parallel datasets including Global Voice, MUSE, News Commentary, Tatoeba\footnote{\url{https://tatoeba.org/en/}}, Talks, WikiMatrix \citep{tiedemann2012parallel,lample2018word,barrault2019findings,reimers2020making,schwenk2021wikimatrix}. We use English as an anchor and transfer NLI abilities to other languages using these paired language data as a bridge. We conduct iterative cycle transfer detailed in the following section.

\subsection{Learning Process}

We use the standard InfoNCE loss \citep{oord2018representation} and make it symmetric as shown in Eq.~\ref{eq:infonce}, in line with CLIP \citep{radford2021learning}. We find that symmetric loss provides extra signals to the learning, resulting in faster and stabler convergence

The loss of a batch $\mathcal{L}$ is defined as:
\begin{equation}
 \small
-\sum_{N} (\log \frac{\exp(s_i\cdot s_i^+) / \tau)}{\sum_{j=0}^{N}\exp(s_i\cdot s_j)/ \tau)} + \log \frac{\exp(s_i^+\cdot s_i) / \tau)}{\sum_{j=0}^{N}\exp(s_i^+ \cdot s_j)/ \tau)}),
\label{eq:infonce}   
\end{equation}
with a batch of N sentences, each $s_i$ normalized. Taking typo-perturbed texts as examples, the model is essentially trained to find, out of all the perturbed examples in the batch, the one corresponding to the original text; and in original examples, the one corresponding to a perturbed text.

\paragraph{Monolingual Learning} We first investigate monolingual learning with English. Motivated by our anisotropy estimates in Section~\ref{subset:poc3} and the established understanding on the uniformity promise of contrastive learning \citep{wang2020understanding,xiao2023isotropy}, we employ a curriculum progressive scheme in displaying the training sets to the model. We start by presenting the easiest examples to the model, facilitating the learning of a) an isotropic representation space 2) robustness towards shape-based perburbations. The intuition here is, if we directly present the supervised training sets to the pretrained model, it has to learn isotropy, shape perception, and reasoning abilities simultaneously with a short training, hindering exploitation of the supervision signals.

Therefore, our training follows a Visual Alignment - Topical Alignment - Reasoning Alignment progression. We facilitate Visual Alignment with shape-based perburbation examples (word shuffling and typos); Topical Alignment is facilitated by independently sampling spans of the same document, from large unsupervised corpus (e.g., Wikipedia), as positive pairs. Last, we inject reasoning abilities into the models, leveraging natural language inference datasets.

\paragraph{Cross-lingual Transfer}

For Cross-lingual Transfer, we first conduct a small-scale experiment to understand the mutual transferability across the languages. We construct and train on each bilingual pairs with training sets of multilingual STS-b, and evaluate the test performance of all languages (2 trained, 8 unseen languages), which gives us a glimpse of the visually-grounded understanding that learning a certain language will bring to other languages.

With these insights, we conduct 3 larger-scale transfers, by constructing parallel datasets of 10, 18, and 59 languages. We construct a mix corpus $\mathcal{M}$ by concatenating English NLI corpus $\mathcal{A}$ and parallel language corpus $\mathcal{P}$. Notably, each language is paired with English in $\mathcal{P}$, because transferring English NLI ability to other languages intuitively has to be achieved by using English as the anchor. Empirically, we find it surprisingly useful to iteratively re-train on $\mathcal{P}$ and $\mathcal{A}$. And after each round that the model is trained on $\mathcal{P}$, it learns better on $\mathcal{A}$, shown by a general STS-b improvement on all languages, until convergence (Section~\ref{sec: cross-lingual transfer results}).

\section{Experiments}

\paragraph{Datasets} For \textit{Visual Alignment}, we use the Wikipedia 1M dataset \citep{gao2021simcse}. We use the TextAttack framework \citep{morris2020textattack} to construct character-level text-perturbed positive pairs and our own implementation for word-level perturbations. For \textit{Topical Alignment}, we use a larger Wikipedia dump, in line with \citet{izacard2021unsupervised} (We construct around 6.5M pairs leveraging the English Wikipedia dump from March 2022), as we speculate topical alignment to be more difficult, with a larger ratio of lexical mismatch between positive pairs. For \textit{Reasoning Alignment}, we use the concatenation of SNLI \citep{bowman2015large} and MNLI \citep{williams2018broad}. This concatenated NLI collection is commonly referred to as all-NLI ($\sim$94k pairs and $\sim$31k entailment pairs). We only use entailment pairs as positive pairs, without leveraging contradiction as hard negative signals.

For cross-lingual transfer, we start from a small scale, using training sets in multilingual STS-b \citep{cer2017semeval,huggingface:dataset:stsb_multi_mt}, with only 5k sentence pairs per language pair, denoted $\mathcal{S}\textsubscript{n}$ with $n$ being the language name. We then use the Ted parallel datasets introduced in \citep{reimers2020making}, with $\sim$30-40k pairs per language. We construct $\mathcal{P}\textsubscript{10}$ and $\mathcal{P}\textsubscript{18}$ with it. With the best configuration, we provide a XL-scale training with the collection of parallel sentences, including Global Voice, MUSE, News Commentary, Tatoeba, Talks, WikiMatrix \citep{tiedemann2012parallel,lample2018word,barrault2019findings,reimers2020making,schwenk2021wikimatrix}, resulting in 59 languages, pushing the limit of cross-lingual performance.

For evaluation, we use the test set of multilingual STS-b \citep{cer2017semeval,huggingface:dataset:stsb_multi_mt}, which include 10 languages from 4 language families as shown throughout the paper. In Appendix~\ref{appendix: IR evaluation}, we further include evaluation of retrieval performance, on Natural Questions \citep{kwiatkowski2019natural}.

\paragraph{Implementation Details} We initialize our models with PIXEL \citep{rust2023language}, which uses a ViT-MAE \citep{he2022masked} backbone and objective, and is pre-trained on the same corpus as BERT \citep{devlin2018bert}, with the training set rendered into images. We use PangoCairo \citep{taylor2004pango} to render texts into images on-the-fly. In most experiments, we train the models for 1 epoch, with a learning rate of 3e-6 in visual alignment step and 3e-5 otherwise, a temperature $\tau$ of 0.05, and a max sequence (patch) length of 64. Mean-pooling is used in all experiments as [cls] token of the base model is under-trained (Appendix~\ref{appendix: pooling mode}). We normalize the final embeddings before loss computation and in inference time, and find it highly beneficial to model convergence, which is consistent with prior work \citep{wang2017normface,wang2020understanding,chen2020simple}. 

\paragraph{Checkpoint Selection} 

Empirically, we find that checkpoints that display good semantic performance in earlier stages do not necessarily provide best potential in later supervised training, exhibiting certain early-phase overfitting. The pattern might also be highly relevant to the performance orthogonality between STS and retrieval tasks \citep{reimers2016task,xiao2023length,muennighoff2023mteb}, with the latter can be highly optimized through our topical alignment step while the former relies on the transferability provided by NLI data. Therefore, we let the models go through all training data in early stages, but take the best checkpoint on STS-b when finally optimized on all-NLI.

\section{Results}

\begin{wraptable}{r}{0.5\textwidth}
\centering
\vspace{-0.4cm}
\begin{threeparttable}
\scalebox{0.64}{
\begin{tabular}{lcccc}\toprule
Metrics$\rightarrow$ &\multicolumn{2}{c}{English} &\multicolumn{2}{c}{OOD Languages (9)} \\\cmidrule{2-5}
Model & STS-b ($\uparrow$) & Ani. ($\downarrow$) & STS-b ($\uparrow$) & Ani. ($\downarrow$) \\
\midrule
PIXEL\textsubscript{vanilla}&16.28 & 0.833 & 22.50 & 0.882\\
{\small + allNLI}& 77.50 & 0.023&48.87&0.529\\
\midrule
{\small + SimCSE\textsubscript{unsup} \citep{gao2021simcse}} &51.51 &0.201& 46.12& 0.469\\
{\small + Character-level Typos} & 52.14 & 0.190 & 47.33 & 0.444 \\
{\quad \small + allNLI}& \underline{77.71} &0.021&49.20&0.508 \\
{\small + Word Conditioned Shuffling} & 56.78 & 0.116 & 43.88 &0.410\\
{\quad \small + allNLI}& \textbf{78.02} & 0.022 & 48.73 & 0.535\\
{\small + Word Random Shuffling} & 60.61 &0.169&49.69&0.450 \\
{\quad \small + allNLI}& 77.40 & 0.022 &48.55 &0.508\\
{\small + Text Pert. Mixure} & 60.73 & 0.156&50.78&0.434\\
{\quad \small + allNLI}& 77.12 & 0.022&49.50&0.480\\
\midrule
{\small + Ensemble} & 59.39&0.145& 51.34& 0.441\\
{\quad \small + allNLI} & 77.78 &0.021&48.96&0.505 \\
{\small + Ensemble + WikiSpan} & 52.74&0.320&40.01&0.552\\
{\quad \small + allNLI} & 77.78 &0.026&\textbf{51.99}&0.525\\

\midrule[0.5pt]\midrule[0.5pt]
BERT\textsubscript{vanilla}&52.59&0.763&43.11&0.904\\
BERT-Flow \citep{li2020sentence}&58.56&-&-&-\\
{\quad \small + allNLI} & 77.08&-&-&-\\
Whitening \citep{su2021whitening}&68.19&-&-&-\\
{\quad \small + allNLI} &78.66&-&-&-\\
SBERT \citep{reimers2019sentence}
&76.98&0.174&44.69&0.712
\\
\bottomrule
\end{tabular}
}
\end{threeparttable}
\vspace{0.2cm}
\caption{Monolingual (Train on English-only) Results.}
\vspace{-1cm}
\label{tab: monolingual results}
\end{wraptable}

In this section, we first present the results of monolingual learning facilitated by our visual-topical-reasoning alignment progression, where we characterize the importance of shape-based perturbation pretraining. We then present results of cross-lingual learning further facilitated by an iterative transfer from English to other languages, where a surprising ``linguistic leapfrogging'' pattern is revealed.

\subsection{Monolingual Learning Results}

Table~\ref{tab: monolingual results} summarizes our monolingual (English-only) learning results. We find a consistent gain by first pre-training on our text perturbation augmentation outlined in our Visual Alignment step, and random span pairs in the Topical Alignment step, before optimizing on NLI datasets. The semantic gain brought by our visual alignment step (Table~\ref{tab: monolingual results} upper part) outperforms training our backbone using SimCSE \citep{gao2021simcse}.
As we analyzed, the unsupervised SimCSE is essentially only forcing a uniform representation distribution of the models by pushing away negative pairs, chiming in with the unifomrity promise of contrastive loss \citep{wang2020understanding}, while our perturbation methods further provide shape-based signals to visually facilitate the alignment promise.
\begin{table*}[!htp]\centering
\scalebox{0.68}{
\begin{tabular}{rc|cccccccccccccccc}
\toprule
\multicolumn{2}{c|}{\multirow{2}{*}{\shortstack{\textbf{Eval ($\rightarrow$)} \\ \textbf{Train ($\downarrow$)}}}} &
\multirow{2}{*}{\textbf{en}} & \multirow{2}{*}{\textbf{de}} & \multirow{2}{*}{\textbf{nl}} & 
\multirow{2}{*}{\textbf{es}} & \multirow{2}{*}{\textbf{fr}} & \multirow{2}{*}{\textbf{it}} & 
\multirow{2}{*}{\textbf{pt}} & \multirow{2}{*}{\textbf{pl}} & \multirow{2}{*}{\textbf{ru}} & 
\multirow{2}{*}{\textbf{zh}} & 
\multicolumn{1}{c}{\textbf{Germanic}} & \multicolumn{1}{c}{\textbf{Romance}} & 
\multicolumn{1}{c}{\textbf{Slavic}} & \multicolumn{1}{c}{\textbf{Sinitic}} & 
\multirow{2}{*}{\textbf{avg.}} \\
& & & & & & & & & & & & \multicolumn{1}{c}{\shortstack{(en, de, nl)}} & \multicolumn{1}{c}{\shortstack{(es, fr, it, pt)}} & \multicolumn{1}{c}{\shortstack{(pl, ru)}} & \multicolumn{1}{c}{(zh)} & \\
\midrule
\multirow{9}{*}{\shortstack{$\mathcal{A}$ \\ + \\ $\mathcal{S}_{n}$}} &\multicolumn{1}{|c|}{\textbf{de}} & 77.40&58.45	&59.44	&58.53	&60.23	&58.95	&58.63	&50.31	&41.35	&23.51	&\textbf{65.09}	&59.09	&45.83	&23.51	&\textbf{54.68}\\
\multicolumn{1}{l|}{}&\textbf{nl} &77.63	&57.45	&\textbf{59.65}	&55.55	&59.21	&57.49	&56.81	&49.26	&38.39	&23.62	&64.91&	57.27	&43.82	&23.62	&53.51\\
\multicolumn{1}{l|}{}&\textbf{es} & 77.48	&58.75	&55.39	&\textbf{59.45}	&59.77	&56.32	&58.22	&48.97	&39.13	&23.06	&63.87	&58.44 &44.05	&23.06	&53.65\\
\multicolumn{1}{l|}{}&\textbf{fr} & 77.50	&\textbf{59.00}	&55.61	&58.67	&\textbf{63.09}	&58.88	&57.83	&49.95	&38.46	&\textbf{24.73}	&64.04	&\textbf{59.62}	&44.21	&\textbf{24.73}	&54.37\\
\multicolumn{1}{l|}{}&\textbf{it} &77.61	&58.35	&55.85	&57.46	&59.96	&\textbf{60.39}	&56.26	&49.42	&38.34	&22.85	&63.93	&58.52	&43.88	&22.85	&53.65\\
\multicolumn{1}{l|}{}&\textbf{pt} &77.41	&58.85	&55.47	&58.26	&59.13	&58.07	&\textbf{61.03}	&48.72	&36.02	&23.70	&63.91	&59.12	&42.37	&23.70	&53.67\\
\multicolumn{1}{l|}{}&\textbf{pl} &76.98	&57.61	&55.17	&54.81	&58.40	&58.34	&56.74	&50.95	&37.30	&20.50	&63.25	&57.07	&44.13	&20.50	&52.68\\
\multicolumn{1}{l|}{}&\textbf{ru} &75.87	&58.30	&55.73	&54.87	&57.69	&57.57	&56.07	&49.97	&43.12	&21.75	&63.30	&56.55	&46.54	&21.75	&53.10\\
\multicolumn{1}{l|}{}&\textbf{zh} & 75.42	&57.05	&55.05	&55.49	&57.29	&57.10	&55.02	&\textbf{51.50}	&\textbf{43.24}	&19.42	&62.50	&56.23	&\textbf{47.37}	&19.42	&52.66\\
\midrule[0.05pt] \midrule[0.05pt] 
\multirow{3}{*}{\shortstack{$\mathcal{P}_{10}$\\ + \\ $\mathcal{A}$}} &\multicolumn{1}{|c|}{\textbf{Iter.1}} &79.46	&62.69	&64.89	&67.47	&68.02	&67.48	&65.97	&64.81	&60.42	&42.17	&69.01	&67.24	&62.61	&42.17	&64.34\\
\multicolumn{1}{l|}{}&\textbf{Iter.2} & 79.85	&64.61	&66.09	&68.48	&69.48	&\colorbox{blue!25}{\makebox[0.7cm]{68.97}}	&67.02	&\colorbox{orange!40}{\makebox[0.7cm]{65.97}}	&62.04	&47.32	&70.19	&68.49	&\colorbox{orange!40}{\makebox[0.7cm]{\textbf{64.00}}}	&47.32	&65.98\\
\multicolumn{1}{l|}{}&\textbf{Iter.3} &79.53	&\colorbox{orange!40}{\makebox[0.7cm]{66.12}}	&66.50	&\colorbox{orange!40}{\makebox[0.7cm]{69.25}}	&\colorbox{orange!40}{\makebox[0.7cm]{69.72}}	&68.90	&\colorbox{orange!40}{\makebox[0.7cm]{67.51}}	&65.87	&61.74	&50.11	&\colorbox{orange!40}{\makebox[0.7cm]{\textbf{70.72}}}	&\colorbox{orange!40}{\makebox[0.7cm]{\textbf{68.85}}}	&63.80	&\textbf{50.11}	&\colorbox{orange!40}{\makebox[0.7cm]{\textbf{66.53}}}\\
\midrule
\multirow{3}{*}{\shortstack{$\mathcal{P}_{18}$\\ + \\ $\mathcal{A}$}} &\multicolumn{1}{|c|}{\textbf{Iter1}} &\colorbox{orange!40}{\makebox[0.7cm]{80.04}}	&61.92	&65.83	&66.64	&68.07	&65.75	&65.01	&62.94	&59.58	&46.58	&69.26	&66.37	&61.26	&46.58	&64.24\\
\multicolumn{1}{l|}{}&\textbf{Iter2} & \colorbox{blue!25}{\makebox[0.7cm]{80.09}}	&63.49	&66.28	&66.87	&68.84	&65.93	&65.52	&65.05	&\colorbox{orange!40}{\makebox[0.7cm]{62.24}}	&49.10	&69.95	&66.79	&63.65	&49.10	&65.34 \\
\multicolumn{1}{l|}{}&\textbf{Iter3} &79.96	&64.59	&\colorbox{orange!40}{\makebox[0.7cm]{66.30}}	&67.46	&69.39	&65.68	&65.84	&65.56	&62.00	&\colorbox{orange!40}{\makebox[0.7cm]{50.57}}	&\textbf{70.28} &\textbf{67.09}	&\textbf{63.78}	&\colorbox{orange!40}{\makebox[0.7cm]{\textbf{50.57}}}	&\textbf{65.74}\\
\midrule
\multicolumn{1}{l|}
{$\mathcal{P}\textsubscript{XL}$}&\textbf{Iter3} &78.79	&\colorbox{blue!25}{\makebox[0.7cm]{68.15}}	&\colorbox{blue!25}{\makebox[0.7cm]{67.60}}	&\colorbox{blue!25}{\makebox[0.7cm]{70.31}}	& \colorbox{blue!25}{\makebox[0.7cm]{70.84}}	&\colorbox{orange!40}{\makebox[0.7cm]{68.92}}	&\colorbox{blue!25}{\makebox[0.7cm]{69.72}}	&\colorbox{blue!25}{\makebox[0.7cm]{66.39}}	&\colorbox{blue!25}{\makebox[0.7cm]{66.76}}	&\colorbox{blue!25}{\makebox[0.7cm]{52.92}}	&\colorbox{blue!25}{\makebox[0.7cm]{71.51}}	&\colorbox{blue!25}{\makebox[0.7cm]{69.95}}	&\colorbox{blue!25}{\makebox[0.7cm]{66.57}}	&\colorbox{blue!25}{\makebox[0.7cm]{52.92}}	&\colorbox{blue!25}{\makebox[0.7cm]{68.04}}\\
\bottomrule
\end{tabular}
}
\caption{Cross-lingual Results. The upper part of the table summarizes the zero-shot transferability by training small-scale on each bilingual pair (\{en, 1 other language\}). The lower part presents large-scale iterative training experiment. At each iteration, we first let the model see the mixture of parallel dataset $\mathcal{P}$ and English NLI $\mathcal{A}$, then let the model revisit English NLI $\mathcal{A}$ again, where we find the leapfrogging pattern - at each round, the models experience a performance boost, as if they have picked up hints across languages to attain newer understanding on others. $\mathcal{P}\textsubscript{10}$,  $\mathcal{P}\textsubscript{18}$ and  $\mathcal{P}\textsubscript{XL}$ have 10, 18, and 59 languages including English, respectively. \colorbox{blue!25}{\makebox[0.9cm]{Number}} denotes best across all settings; \colorbox{orange!40}{\makebox[0.8cm]{number}} denotes second best. \textbf{Bold} denotes the best in its own setting.}
\label{tab: cross-lingual results}
\end{table*}




We additionally find that taking an ensemble of all (4) unsupervised checkpoints improve the model's OOD generalization. This is further enhanced by topical alignment (denoted WikiSpan). We find pretraining on WikiSpan is essential to trigger exceptional zero-shot OOD generalization when fine-tuning on allNLI, and also provides a good initialization for downstream cross-lingual training.

We further present several most-recognized NLP baselines initialized with BERT. Our performance is on-par with methods such as SBERT \citep{reimers2019sentence}, BERT-whitening \citep{su2021whitening}, BERT-Flow \citep{li2020sentence}. We highlight that we achieve such performance with a vanilla checkpoint with subpar semantic performance, characterized by the fact that vanilla PIXEL is 36 absolute points behind vanilla BERT on sentence-level semantics. This limitation is evident by that the unsupervised SimCSE \citep{gao2021simcse} can achieve over 76.85 with BERT and 80.22 with RoBERTa, but only 51.51 with PIXEL with our implementation. We envision the advancement of the line of work pioneered by our method will be greatly benefited from better pretrained checkpoints in the future.

We refer readers to Appendix~\ref{appendix: IR evaluation} for retrieval evaluation.

\subsection{Cross-lingual Transfer Results}
\label{sec: cross-lingual transfer results}

Table~\ref{tab: cross-lingual results} presents our findings on Cross-lingual Transfer.

The upper part of Table~\ref{tab: cross-lingual results} presents the results of the 9 small-scale transfer settings, created by concatenating English NLI dataset $\mathcal{A}$ with STS-b bilingual pair $\mathcal{S}_n$, where $n$ is each of the language from \{en, de, nl, es, fr, it, pt, pl, ru, zh\}. When $n=$ en, there isn't alignment dataset, and the result falls back to only training on $\mathcal{A}$ (thus monolingual results). Through this part of the results, the findings are summarized as follows: 1) In general, training on $\mathcal{S}_n$ provides zero-shot generalization on languages from the same language family. For instance, training on $\mathcal{S}\textsubscript{German}$ (de) provides good STS-b performance on Dutch. And the same applies to languages among the Romance family.

2) The only outlier is $\mathcal{S}\textsubscript{Chinese}$ (zh), which surprisingly hurts its own semantic performance, but provides good transfer for Russian and Portuguese, which we hypothesize to be due to its shape's being too out-of-distribution in pre-training, and thus being extremely unstable when getting aligned using few paired examples.

The lower part presents our iterative training going back and forth parallel datasets and English NLI. The number reported for each iteration is after optimized on NLI again after ``cross-lingual exploration''. We have three settings, $\mathcal{P}\textsubscript{10}$, $\mathcal{P}\textsubscript{18}$ and $\mathcal{P}\textsubscript{XL}$, with 10, 18 and 59 languages respectively.

\begin{figure*}
    \centering
    \includegraphics[width=1\textwidth]{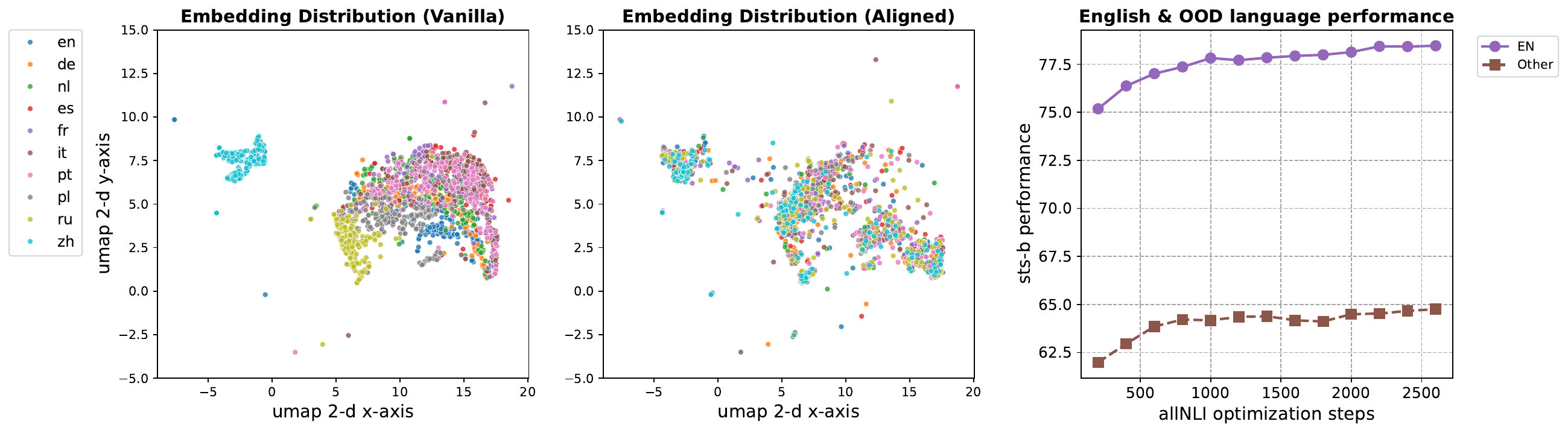}
    \caption{\textbf{Left 1-2:} Embedding Distribution of the vanilla model and model after 3 rounds of iterative alignment. \textbf{Left 3:} English and OOD language performance in the final optimization of allNLI. After alignment, English and other languages present a bonding effect.}
    \label{fig:distribution}
\end{figure*}

\paragraph{Linguistic Leapfrogging} We find a ``leapfrogging''-like pattern presents in our cross-lingual learning process. In our monolingual training, the model's English sts-b performance is capped at around 78 regardless how much extra pre-training we do before fine-tuning it on English all-NLI, which we attributed largely to the inadequacy of the pretrained checkpoint as discussed earlier. However, after the model is further trained on English-Other language pairs, and goes back to revisit all-NLI, its English sts-b performance surprisingly surpasses the previous ceiling, achieving a performance surpassing 80. This clearly reveals that models have learned to infer on the semantics of English through finding helpful ``hints'' when trained on other languages.

\paragraph{Visualizing Language Alignment} In Figure~\ref{fig:distribution}, we visualize the alignment across languages at the end of our iterative training. Going from occupying distinctive subspaces in vanilla model (Fig~\ref{fig:distribution}, left 1), all languages overlap in a larger space after contrastive alignment training (Fig~\ref{fig:distribution} left 2. This alignment enables ``bonding''  across languages: In (Fig~\ref{fig:distribution}, left 3), we visualize the mutual growth of all language semantics in the last optimization round on English all-NLI. Clearly, the subspaces are well-aligned in parallel training, so when optimizing on English only, all languages ``bond'' together and optimize without being directly optimized.

\section{Analysis}

\subsection{Traditional Visual Augmentation v.s.\\ Visually-grounded Text Augmentation}

\begin{wraptable}{r}{0.5\textwidth}
\centering
\vspace{-2cm}
\scalebox{0.75}{
\begin{tabular}{lrr}\toprule
Model & STS-b ($\uparrow$) & Anisotropy ($\downarrow$) \\\midrule
PIXEL\textsubscript{vanilla}&16.28 & 0.833\\
{\small + Cropping} & 4.07 & 0.187 \\
{\quad \small + allNLI}& 77.11 & 0.216\\
{\small + Horizontal Flip} & 5.97 & 0.218\\
{\quad \small + allNLI}& 77.19 & 0.211\\
{\small + Vertical Flip} & 24.75 & 0.650\\
{\quad \small + allNLI}& 76.99&  0.213\\
{\small + Gaussian Blurring} & 23.59 & 0.560\\
{\quad \small + allNLI}&76.98   &0.220\\
\midrule
{\small + Aug. Mixture} &4.07  & 0.187\\
{\quad \small + allNLI}&77.11 & 0.216\\
\midrule[0.5pt]\midrule[0.5pt]
Pixel-Linguist\textsubscript{\textbf{\texttt{en}}}-best & \multicolumn{2}{c}{16.28 (vanilla) $\rightarrow$ 78.02 (ours)}\\
Pixel-Linguist\textsubscript{\textbf{\texttt{all}}}-best & \multicolumn{2}{c}{16.28 (vanilla) $\rightarrow$ 80.09 (ours)}\\
\bottomrule
\end{tabular}
}
\vspace{0.2cm}
\caption{Unsupervised vision augmentation performance.}
\vspace{-1cm}
\label{tab: vis aug}
\end{wraptable}

\label{ablation: tradition vis aug}
With the proven performance gain of our visually-grounded text-based augmentation, it is a natural question whether standard visual augmentation methods could replace our methods. We experiment with four representative visual augmentation methods, cropping, horizontal flip, vertical flip, and Gaussian blurring \citep{chen2020simple}, and the mixture of them (Table~\ref{tab: vis aug}), and show that traditional methods lead to a severe semantic collapse, and it is not recoverable with NLI, largely falling behind our textual perturbations.

\subsection{Does the framework apply to tokenization-based models?}

\begin{table}[tp]
\centering
\scalebox{0.8}{
\begin{tabular}{lcccc}\toprule
Model & BERT & mBERT & CANINE & ByT5\\\midrule
{\small Vanilla} &52.59& 50.98 & 48.36 &22.12\\
{\small + allNLI} & 80.59 & 78.01 & 68.29 & 67.43 \\
{\small + Character-level Typos} & 62.16 & 61.88 &50.71&47.01\\
{\quad \small + allNLI}&80.93 &78.14&67.84& 67.43\\
{\small + Word Conditioned Shuffling} & 71.99 &68.61&36.91& 39.65\\
{\quad \small + allNLI}&81.24 &78.72&68.25& 68.63\\
{\small + Word Random Shuffling} & 73.81 &66.54 &43.33&34.63\\
{\quad \small + allNLI}& 81.50& 78.34&69.25&70.05\\
{\small + Text Pert. Mixture} & 46.78&46.58&47.46& 46.08\\
{\quad \small + allNLI}& 81.12&78.21 &69.26&69.64\\

\midrule[0.5pt]\midrule[0.5pt]
Pixel-Linguist\textsubscript{\textbf{\texttt{en}}}-best & \multicolumn{4}{c}{16.28 (vanilla) $\rightarrow$ 78.02 (ours)}\\
Pixel-Linguist\textsubscript{\textbf{\texttt{all}}}-best & \multicolumn{4}{c}{16.28 (vanilla) $\rightarrow$ 80.09 (ours)}\\
\bottomrule
\end{tabular}
}
\vspace{0.2cm}
\caption{LM performance with our methods.}
\vspace{-0.5cm}
\label{tab: bert}
\end{table}

We experiment with whether our perturbation-based methods apply to language models using tokenization. We consider subword-level, character-level and byte-level language models, including BERT \citep{devlin2018bert}, mBERT \citep{devlin2018bert}, Canine \citep{clark2022canine} and ByT5 \citep{xue2022byt5}, providing a complete picture of the pattern of our methods applied to LMs.

Table~\ref{tab: bert} summarizes the results (see Appendix~\ref{appendix: LM implementation} for technical details). Unexpectedly, our novel unsupervised methods do work for conventional language models, showing the universality of our method. While we believe in pixel models, we do encourage future work to understand this pattern.

\section{Conclusion}

In this work, we proposed a novel pixel-based sentence representation learning framework, employing visually grounded textual perturbation as an unsupervised contrastive pairs construction strategy. We proved the superiority of the framework in its effectiveness and robustness. We uncovered a ``leapfrogging'' pattern, where learning across languages enhances the model's understanding on each language individually. These patterns validate that modeling textual semantics in the pixel space, utilizing shape-based information inherent in text, is a promising path for learning stronger and more human-like sentence encoders.

\section*{Impact Statements}

This research primarily focuses on exploring an innovative way for learning natural language semantics, providing an alternative avenue diverging from tokenization-based LMs for the research community. We anticipate minimal societal impact of this work.  Though our method is partly inspired by human cognition, it is without immediate applications to human-interactive or real-world environments. Should future developments extend our methods to applications involving direct human interaction or decision-making systems, it will be essential to carefully consider and address potential ethical, safety, and fairness concerns.

Additionally, we show that cross-lingual transfer and semantics performance can be achieved with relative small-scale vision models compared to nowadays LLMs. Through this approach, we aim to contribute to the development of more efficient computational models, potentially mitigating their environmental impact over time.

Lastly, we envision that the zero-shot transferability presented in our experiments might provide positive impact on understanding and protecting low-resource languages. For this purpose, however, we advocate for these methods to be employed alongside human verification, particularly confirmation by native 
speakers, ensuring accuracy and respect for linguistic diversity.

\bibliography{arxiv_main}
\bibliographystyle{apalike}

\newpage
\appendix
\onecolumn

\section{Information Retrieval Results}
\label{appendix: IR evaluation}

We implement our evaluation of information retrieval with the BEIR framework \citep{thakur2021beir}. We wrap the pixel encoders into the framework.

The models are mainly evaluated on Natural Questions \citep{kwiatkowski2019natural}, considering it is general-domain task, it is large-scale, and commonly used to evaluate retriever models. The performance is quantified by nDCG@n, and recall@n, which respectively concern (\textbf{nDCG@n:}) how well the top $n$ retrieved documents are ranked, and (\textbf{recall@n:}) if the ground-truth documents are even successfully retrieved into top $n$.

\begin{table}[h]
\centering
\scalebox{0.9}{
\begin{tabular}{lcccc}\toprule
Method & nDCG@1 & nDCG@10 & recall@1 & recall@10\\\midrule

{\small + ensemble} &0.012& 0.026 & 0.011 & 0.043\\
{\small + ensemble + wikispan} & 0.002 & 0.003 & 0.002 & 0.004 \\
\midrule
{\small + allNLI} &0.076& 0.143 & 0.068 & 0.224\\
{\small + ensemble + allNLI} & \underline{0.082} & \underline{0.148} & \underline{0.075} & \underline{0.231} \\
{\small + ensemble + wikispan + allNLI} & \textbf{0.091} & \textbf{0.175} & \textbf{0.078} & \textbf{0.283} \\
\bottomrule
\end{tabular}
}
\vspace{0.2cm}
\caption{Information Retrieval Results on Natural Question}
\label{tab: IR results}
\end{table}

The results are presented in Table~\ref{tab: IR results}. The findings are interesting: 1) without ending in allNLI, training wikispan after ensemble actually degrades the model's performance. This is in contrast to findings of \citet{izacard2021unsupervised} in NLP, who find that training with Wikipedia spans provides strong IR performance. 2) However, the advantage of Wikispan is immediately pronounced when the model is further trained with allNLI, outperforming other training order. Therefore, it can be concluded that training on Wikispan does enhance retrieval performance, but with pixel models, Wikispan acts more like a pretraining task, and its advantages need to be triggered with supervised data like allNLI, or supervised IR datasets.

\section{Implementation of LM ablation}
\label{appendix: LM implementation}
We use mean-pooling to attain sentence representation for all language model ablation experiments, aligning with pixel experiments. For ByT5 \citep{xue2022byt5}, we use the T5 encoder only, aligning with previous works in sentence-level semantics.

\section{Pooling Mode}
\label{appendix: pooling mode}
As mentioned in implementation details, we use mean-pooling (instead of \texttt{[cls]} pooling). This is because we find that the \texttt{[cls]} token in PIXEL, though present, is not leveraged in the model's pretraining objectives, and thus present also perfect anisotropy (Table~\ref{tab: appendix pooling mode}), with two randomly-sampled sentences presenting a cosine similarity of 0.999. By contrast, BERT's NSP objective in pretraining more or less pushes \texttt{[cls]} embeddings of different sentences apart. Based on these behaviors, we choose to apply mean-pooling to PIXEL.

\begin{table}[h]
\centering
\scalebox{0.9}{
\begin{tabular}{lrrrrr}\toprule
Pooling$\rightarrow$ &\multicolumn{2}{c}{[CLS]-pooling} &\multicolumn{2}{c}{Mean-pooling} \\\cmidrule{2-5}
Language$\downarrow$ &BERT &PIXEL &BERT &PIXEL \\\midrule
en &\textbf{0.926} &0.999 &\textbf{0.763} &0.833 \\
de &\textbf{0.971} & 0.999 & 0.895 & \textbf{0.877}\\
nl &\textbf{0.965} &0.999 &0.895 &\textbf{0.867} \\
\midrule
es &\textbf{0.974} &0.999 &0.901 &\textbf{0.894} \\
fr &\textbf{0.970} &0.999 &0.893 &\textbf{0.879} \\
it &\textbf{0.970} &0.999 &0.900 &\textbf{0.884} \\
pt &\textbf{0.968} &0.999 &0.891 &\textbf{0.874} \\
\midrule
pl &\textbf{0.975} &0.999 &0.914 &\textbf{0.866} \\
ru &\textbf{0.977} &0.999 &0.942 &\textbf{0.917} \\
\midrule
zh &\textbf{0.963} &0.999 &0.909 &\textbf{0.880} \\
\bottomrule
\end{tabular}
}
\vspace{0.2cm}
\caption{Anisotropy Estimates ($\downarrow$ the better) with different pooling modes.}
\label{tab: appendix pooling mode}
\end{table}

\end{document}